\documentclass[10pt,twocolumn,letterpaper]{article}

\usepackage[pagenumbers]{cvpr}

\usepackage[T1]{fontenc}
\usepackage{newunicodechar}
\newunicodechar{ł}{\l}
\usepackage{booktabs}
\usepackage{colortbl}
\usepackage{multirow}
\usepackage{array}
\usepackage{tabularx}
\usepackage{xspace}
\usepackage{amsmath}
\usepackage{amssymb}
\usepackage{pifont}
\usepackage{enumitem}
\usepackage{algorithm}
\usepackage{algpseudocode}
\newcommand{\ours}{Atomic Policy\xspace}
\newcommand{\phase}[1]{\textsf{#1}}

\newcommand{\cmark}{\ding{51}}
\newcommand{\xmark}{\ding{55}}
\newcommand{\sd}[1]{\,{\scriptsize\textcolor{black!55}{$\pm$\,#1}}}

\definecolor{cvprblue}{rgb}{0.21,0.49,0.74}
\usepackage[pagebackref,breaklinks,colorlinks,allcolors=cvprblue]{hyperref}

\def\confName{CVPR}
\def\confYear{2027}

\title{Atomic Policy: Composing Atomic Manipulation Skills with Agentic Reasoning}

\author{Kaidi Zhang$^*$ \quad Pengyuan Guo$^*$ \quad Zhengtong Xu$^*$ \quad Caiyi Zhang \quad Huayi Gou\\
Guanlin Huang \quad Aaron Lim \quad Anas Rafei \quad Yu She\\[2pt]
Purdue University\\[2pt]
{\small $^*$Equal contribution}
}

\makeatletter
\apptocmd{\@maketitle}{%
  \par
  \begin{minipage}{\textwidth}
    \captionsetup{type=figure}
    \centering
    \includegraphics[width=0.98\linewidth]{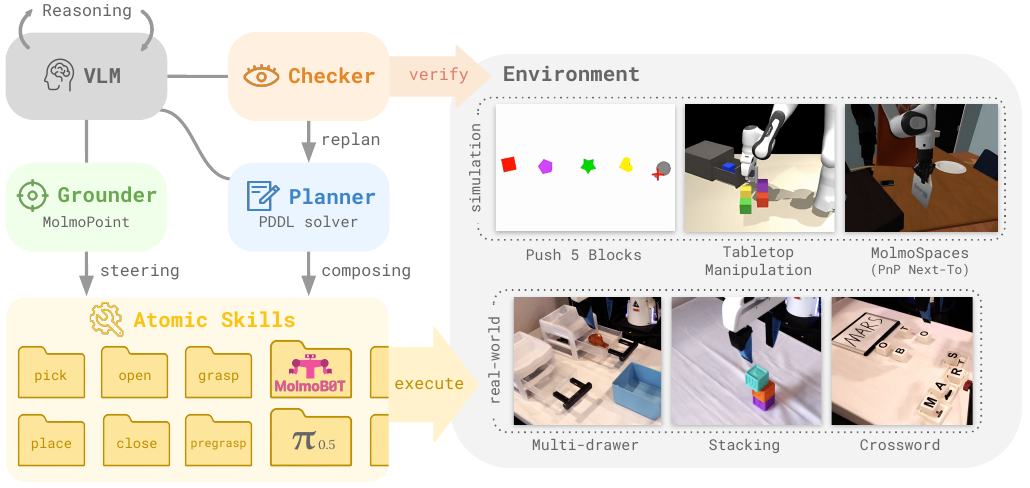}
    \caption{\textbf{Atomic Policy} integrates high-level reasoning with learned atomic skills for solving complex tasks. A VLM interprets user intent from language and scene observations, while a PDDL solver composes atomic skills into executable plans. Language instructions and visual pointing steer low-level execution, and a separate VLM checker verifies phase completion and triggers replanning when necessary.}
    \label{fig:teaser}
  \end{minipage}
  \par\medskip
}{}{%
  \PackageError{AtomicPolicy}{Could not insert the teaser into the title block}{}
} 
\makeatother

\begin{document}
\maketitle
\begin{abstract}
Long-horizon manipulation requires robots to translate task intent into physical interactions as the scene changes.
Vision-language models (VLMs) offer semantic and spatial reasoning, while visuomotor policies provide fine-grained control. Yet ambiguous goals and dependencies between actions make their integration challenging.
We introduce Atomic Policy, a framework that composes learned manipulation skills through grounded steering and execution feedback.
A VLM estimates the symbolic state and goal, and a PDDL solver constructs a skill plan.
An agentic harness steers each skill through language, visual pointing, or both, and verifies its outcome to guide phase transitions and replanning.
A common phase-level interface supports specialized policies and generalist action models.
Experiments in simulation and on a real robot show how Atomic Policy improves task following and supports recovery in complex, long-horizon manipulation.
\iftoggle{cvprfinal}{Project page: \href{https://aaronzkd.github.io/atomic-policy/}{\texttt{aaronzkd.github.io/atomic-policy}}}{}
\end{abstract}
    
\section{Introduction}

Long-horizon manipulation requires a robot to pursue its task goal while reasoning about a scene that changes with every interaction.
Instructions may leave intermediate steps implicit, and an unsuccessful action can alter what is feasible next.
Vision-language models (VLMs) provide a basis for this reasoning, enabling robotic agents to interpret instructions and coordinate perception, planning, and control~\cite{fu2026capx,chen2026volo,guo2026planar}.
Perceptual and geometric tools help these agents resolve spatial relationships and ground ambiguous or underspecified instructions in the scene~\cite{chen2026spacetools}.
Visual feedback then allows them to monitor progress and revise decisions as execution unfolds~\cite{chen2026volo,gemini2025robotics}.

Translating these decisions into reliable manipulation remains difficult.
The physical interactions available to an agent are constrained by its execution interface~\cite{fu2026capx,zhang2026playful}: identifying the correct object and operation does not specify the motion needed to complete the interaction.
Analytic primitives can struggle with complex grasps, constrained placement, and articulated-object interaction~\cite{zhang2026harness}.

Imitation learning provides fine-grained motor behaviors learned from demonstrations~\cite{zhao2023act,chi2023diffusion}.
Vision-language-action models (VLAs) extend this capability to language-conditioned control, demonstrating semantic generalization and extended task execution in unfamiliar settings~\cite{black2025pi05,gemini2025robotics}.
Reliable execution nevertheless remains challenging when a familiar behavior must be redirected toward a different target or composed with other skills in a new sequence~\cite{chen2026volo,zhang2026harness}.

Recent efforts address skill reuse and coordination through procedural knowledge and code-based skills~\cite{lu2026aspire,he2026skill2real}, orchestration of learned motor policies~\cite{li2026roboclaw,zhang2026robocop,chen2026volo,zhang2026harness}, and skill transitions guided by world models~\cite{zhang2026wap}.
We focus on how task reasoning can steer these motor behaviors while remaining responsive to their outcomes.
The agent must specify the intended interaction and its target, then determine whether the resulting state supports the next step.
Grounded steering and verification therefore become part of skill composition.

We introduce \textit{Atomic Policy} (Fig.~\ref{fig:teaser}), a framework that separates local manipulation skills from task-level reasoning and composition.
Each atomic skill is a reusable behavior with a phase-level completion condition, implemented by either a dedicated visuomotor policy or a shared language-conditioned model.
A VLM-based agentic harness estimates the symbolic state and goal, and a PDDL solver composes the skills according to their preconditions and effects.
The harness steers each phase through language, visual pointing, or both: language specifies the interaction and its referents, while points make spatial targets explicit.
A VLM checker verifies the outcome and determines whether to continue, advance, or replan from the current scene.
This common interface allows task reasoning to redirect and compose skills while learned policies generate local motion, supporting recovery as execution unfolds.

\begin{figure*}[t]
  \centering
  \includegraphics[width=\textwidth]{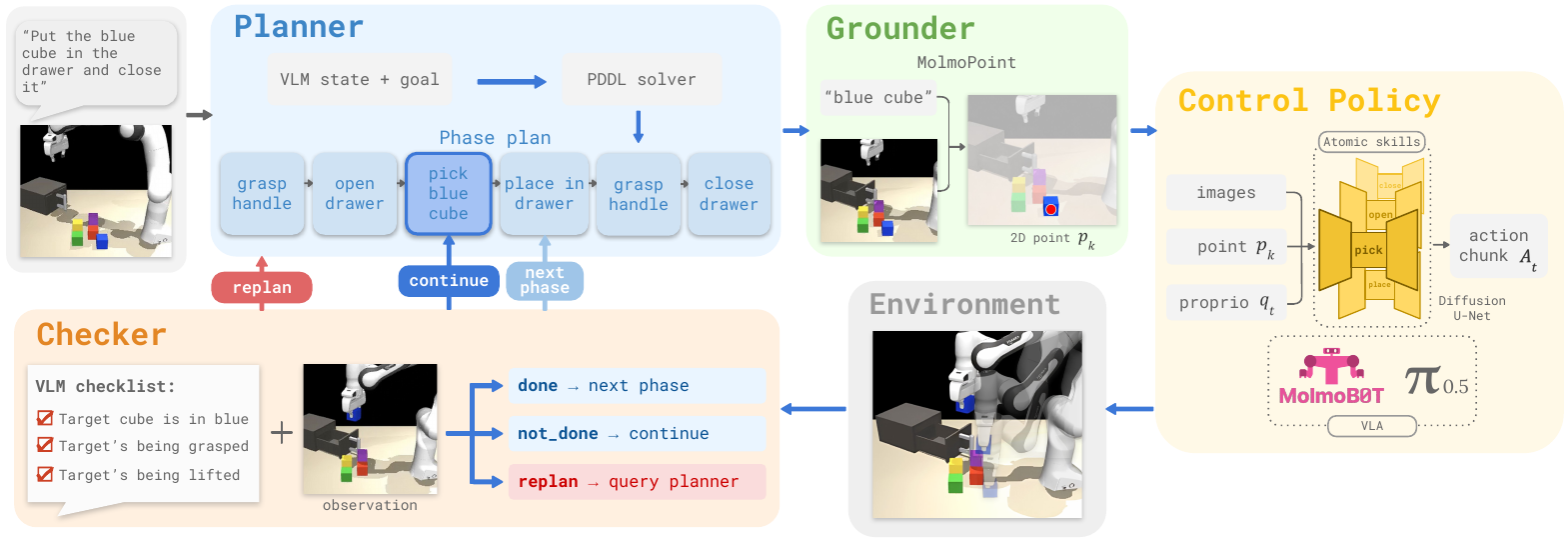}
  \caption{\textbf{Atomic Policy architecture}, shown with point-steered skills for Tabletop Manipulation.
  The \emph{planner} combines VLM state and goal estimation with PDDL planning.
  For the active \emph{pick blue cube} phase, the \emph{grounder} (MolmoPoint) supplies a 2D image point $p_k$.
  The selected skill generates an action chunk $A_t$ using a diffusion U-Net conditioned on images, proprioception $q_t$, and $p_k$.
  The VLM \emph{checker} evaluates the outcome against a phase-specific checklist: \texttt{done} advances the plan, \texttt{not\_done} continues the phase, and \texttt{replan} requests a new plan.
  Blue arrows indicate execution and phase transitions; the red arrow indicates replanning.}
  \label{fig:overview}
\end{figure*}

Our contributions are threefold.

(1) We introduce Atomic Policy, a framework for complex, long-horizon manipulation that connects task-level reasoning to learned skill execution through grounded steering and verification.

(2) We formulate point and language steering as complementary interfaces for skill composition. A common phase-level abstraction supports specialized visuomotor policies and generalist action models within the same planning and execution loop.

(3) We evaluate the framework in simulation and on a real robot across tasks requiring dependent skill sequences and interpretation of ambiguous or underspecified goals. Comparisons, ablations, and failure analysis examine how steering and replanning contribute to task following and recovery.

\section{Related Work}
\label{sec:related}

\paragraph{Agentic Robotics.}
Frontier multimodal models provide reasoning and tool-use capabilities~\cite{comanici2025gemini25}.
Robot agents use these capabilities to call learned policies as tools~\cite{chen2026volo,li2026roboclaw,zhang2026robocop} or generate executable robot programs~\cite{fu2026capx,shi2026maestro,xie2026pusht}.
Harness VLA~\cite{zhang2026harness} combines a frozen VLA contact primitive with analytic actions.
Other systems co-design physical tools and action plans~\cite{gao2026vlmgineer} or learn to use tools for spatial reasoning~\cite{chen2026spacetools}.
Agents also acquire reusable skills or refine programs through practice~\cite{lu2026aspire,zhang2026playful,he2026skill2real,chen2026gap}, and improve policies or execution through real-world trials~\cite{xiao2026enpire}.
Agent-generated demonstrations and program annotations also support policy learning~\cite{shen2026embodiedswe,liu2026aw0}.
Benchmarks test automation~\cite{xie2026vabar}, manipulation for evidence gathering~\cite{bhat2026mgvqa}, and memory~\cite{wang2026memoreval}.
Plans can also be evaluated using world and video models~\cite{zhang2026wap,fu2026novaplan}.
Neuro-symbolic representations further structure task reasoning~\cite{mao2025neurosymbolic}.
VLMs support task planning~\cite{hu2023vila} and can be coupled with PDDL-based or task-and-motion planning~\cite{zhang2024dkprompt,yang2024vlmtamp,mishra2026tampagent}.
PLanAR~\cite{guo2026planar} uses symbolic action schemas for verification and replanning, with grasp-pose generation and parameterized primitives for execution.
Other work learns composable skills from unsegmented demonstrations~\cite{shao2026symskill,nie2026stack} or composes short-horizon plan distributions through search~\cite{mishra2026cdgs}.
VLM-based systems also support visual failure detection and execution monitoring~\cite{duan2024aha,zhou2024cam}.
\ours formulates skill execution through a common phase interface, coupling the selected behavior and target to a completion condition.
This supports dedicated policies and shared action models within symbolic planning and verification.

\paragraph{Generalist Policies and Grounding Interfaces.}
Imitation learning acquires visuomotor skills from demonstrations~\cite{zhao2023act,chi2023diffusion}.
Generalist policies scale imitation learning to large robot datasets~\cite{octo2024,kim2024openvla,black2024pi0}.
Vision-language-action models support language-conditioned robot control~\cite{black2025pi05,bjorck2025gr00t,gemini2025roboticsvla,zhang2026tacvla}.
Simulation training enables zero-shot transfer to real robots~\cite{molmobot2026,xue2025doorman}, while RoboArena evaluates generalist policies across institutions~\cite{atreya2025roboarena}.
When conditioned only on a full-task prompt, a policy must infer the current subtask and target during execution.
Intermediate reasoning takes the form of language motions~\cite{belkhale2024rth}, embodied reasoning steps~\cite{zawalski2024ecot}, or subtask instructions from a high-level model~\cite{shi2025hirobot,gemini2025robotics}.
Embodied reasoning models integrate localization, spatial reasoning, and long-horizon planning~\cite{bjorck2026vesta}.
Pointing VLMs ground language in image locations, including objects and free-space regions~\cite{deitke2024molmo,yuan2024robopoint,zhou2025roborefer,molmopoint2026}.
PointArena~\cite{cheng2025pointarena} benchmarks this capability.
Visual spatial cues guide action selection, motion planning, or optimization~\cite{liu2024moka,nasiriany2024pivot,huang2024rekep,huang2024copa}.
Trajectory sketches, visual traces, and 3D contact points also condition learned policies~\cite{gu2023rttrajectory,zheng2024tracevla,li2025hamster,cui2026cap}.
CAP composes contact-conditioned skills through VLM tool calling, verification, and retries.
\ours supports dedicated skill policies and generalist action models through language, 2D target points, or both, with symbolic planning and phase-level verification.
For point steering, a pretrained grounder requires neither depth input nor task-specific fine-tuning.

\section{Method}
\label{sec:method}

\ours composes learned manipulation behaviors through a phase-level interface that specifies the active skill, its target, and the evidence required for completion.
A dedicated policy or shared action model executes each phase, steered by language instructions, image points, or both.
The harness grounds task intent, composes phases, and uses visual feedback to continue, advance, or replan.
\Cref{fig:overview} illustrates the framework through a drawer-manipulation task.

\subsection{General Formulation and Closed-Loop Execution}
\label{sec:method_formulation}

The robot receives a language instruction $\ell$ and observes $o_t = (I_t^{1},\dots,I_t^{M}, q_t)$ at control step $t$, where the images come from $M$ cameras and $q_t$ is the proprioceptive state.
A plan $\mathcal{P}_t$ consists of $K$ phases $P_k = (s_k, g_k)$, $k = 1, \dots, K$, where $s_k \in \mathcal{S}$ identifies an atomic skill and $g_k$ describes its target, such as ``the blue cube'' or ``the third target circle''.
We use $g_k=\varnothing$ when a phase requires no separate target, as in opening an already-grasped drawer.
An atomic skill is a reusable manipulation behavior with a phase-level completion condition.
It serves as a unit of task composition and may span multiple control steps and policy queries.
Each skill can be implemented by a dedicated policy or by a shared policy conditioned on the phase instruction.

\label{sec:method_plan}
Each task family has a PDDL domain~\cite{mcdermott1998pddl}, denoted by $\mathcal{D}$, whose operators describe the available skills through their preconditions and effects.
Given this domain, the instruction, and the current image, the VLM estimates the ground predicates that hold in the scene, $\hat{x}_t$, and the goal predicates implied by the instruction, $G$.
Formally, $(\hat{x}_t,G) = \operatorname{VLM}(\mathcal{D},\ell,I_t^{1})$, and the symbolic solver constructs the phase sequence $\mathcal{P}_t = \operatorname{Solve}(\mathcal{D},\hat{x}_t,G)$.
The solver searches over applicable operators and maps the resulting sequence to the phases $P_k$.
The plan is valid with respect to the estimated state and domain; visual verification checks whether its intended effects occur during execution.

Consider the instruction ``Put the blue cube in the drawer and close it'' in \cref{fig:overview}.
With the drawer initially closed and the cube on the table, the plan is the following sequence of $(s_k,g_k)$ pairs:
\begin{equation}
  \label{eq:plan_example}
  \begin{aligned}
    & (\phase{grasp},\text{drawer handle}) \to (\phase{open},\varnothing) \\
    & \quad\to (\phase{pick},\text{blue cube}) \to (\phase{place},\text{drawer}) \\
    & \quad\to (\phase{grasp},\text{drawer handle}) \to (\phase{close},\varnothing).
  \end{aligned}
\end{equation}
The plan respects dependencies such as grasping before opening and picking before placement.
After placement, the \phase{grasp} skill is reused to hold the handle for closing.
The \phase{open} and \phase{close} phases act on the already-grasped drawer without a separate point target.

For the active phase, the harness forms a steering input $u_k$ from the skill and target description and invokes the policy as $A_t \sim \pi_{s_k}(\,\cdot \mid o_t, u_k)$.
Here $A_t$ is an action chunk, and $u_k$ is a target point, a phase instruction, or both.
For \phase{pick}, a dedicated picking policy receives the blue cube's image point, while a language-conditioned policy receives an instruction to pick up that cube.
The policy generates local motion through repeated queries from the latest observations until the checker advances the phase or requests replanning.

\label{sec:method_check}
A VLM checker periodically evaluates the current phase against a task-specific checklist using visual observations and robot state.
It returns \texttt{not\_done} to continue execution, \texttt{done} to advance to the next phase, or \texttt{replan} when the remaining plan needs revision.
For example, advancing from \phase{pick} to \phase{place} requires visual evidence that the blue cube is grasped and lifted (\cref{fig:overview}).
On \texttt{replan} or a phase timeout, the planner re-estimates the symbolic state and solves from the latest observation.
If the drawer is already open, the new plan can skip the opening sequence and proceed with the remaining phases.
Execution resumes from the current observation with updated steering inputs, closing the planning and verification loop.
The planner, grounder, and checker require no training on task-specific robot demonstrations.
Implementation details and execution pseudocode are provided in \cref{sec:appendix_method}.

\subsection{Point Grounding and Steering}
\label{sec:method_ground}

Point steering makes the spatial target of a skill explicit.
For each target-dependent phase, the harness turns $g_k$ into a pointing query, such as ``Point to the blue cube''.
MolmoPoint-8B~\cite{molmopoint2026} grounds this query in the current primary-camera image and returns a 2D point $p_k \in [0,1]^2$ in normalized image coordinates.
The point is refreshed during execution so that the policy can follow changes in the target's image location.

In the drawer example, grounding identifies the handle for \phase{grasp}, the blue cube for \phase{pick}, and the drawer as the destination for \phase{place}, providing the spatial reference needed for each interaction.

The selected policy receives $u_k=p_k$ along with its visual and proprioceptive observations.
Skill selection specifies the behavior, and the point identifies its target; the policy determines the approach and interaction from its observations.
This lets the harness resolve semantic intent and direct a skill toward a named target without passing the full task instruction to that skill.
For phases without a separate target, the point input is omitted.
Our specialized skill bank learns these behaviors from phase-segmented demonstrations; its architecture and training details are given in \cref{sec:method_policy}.

\subsection{Language Grounding and Steering}
\label{sec:method_language}
\label{sec:method_vla}

For language steering, the harness expresses the selected skill and its resolved target as a short phase instruction.
We use phase-specific templates; for example, ``grasp the \{object\} and lift it'' becomes ``grasp the blue cube and lift it''.
The resulting phase instruction $\ell_k$ provides the steering input $u_k=\ell_k$ to a language-conditioned policy.
The harness specifies the immediate operation and its referent, while the policy localizes the named target in its observations and generates the motion.
Phase transitions and replanning follow the same verification loop as point steering.

Vision-language-action models provide a natural implementation of this interface because they already map language, images, and robot state to actions.
A shared model can therefore execute different atomic skills through phase-specific prompts.
We instantiate language steering with $\pi_{0.5}$, using phase prompts without points because its native interface does not support point conditioning.

Existing policies also support combined steering: MolmoBot~\cite{molmobot2026} accepts language and point inputs, while GR00T 2~\cite{nvidia2026gr00t2} uses language instructions with visual grasp and release markers.
\section{Experiments}
\label{sec:experiments}

\begin{figure*}[t]
  \centering
  \includegraphics[width=\textwidth]{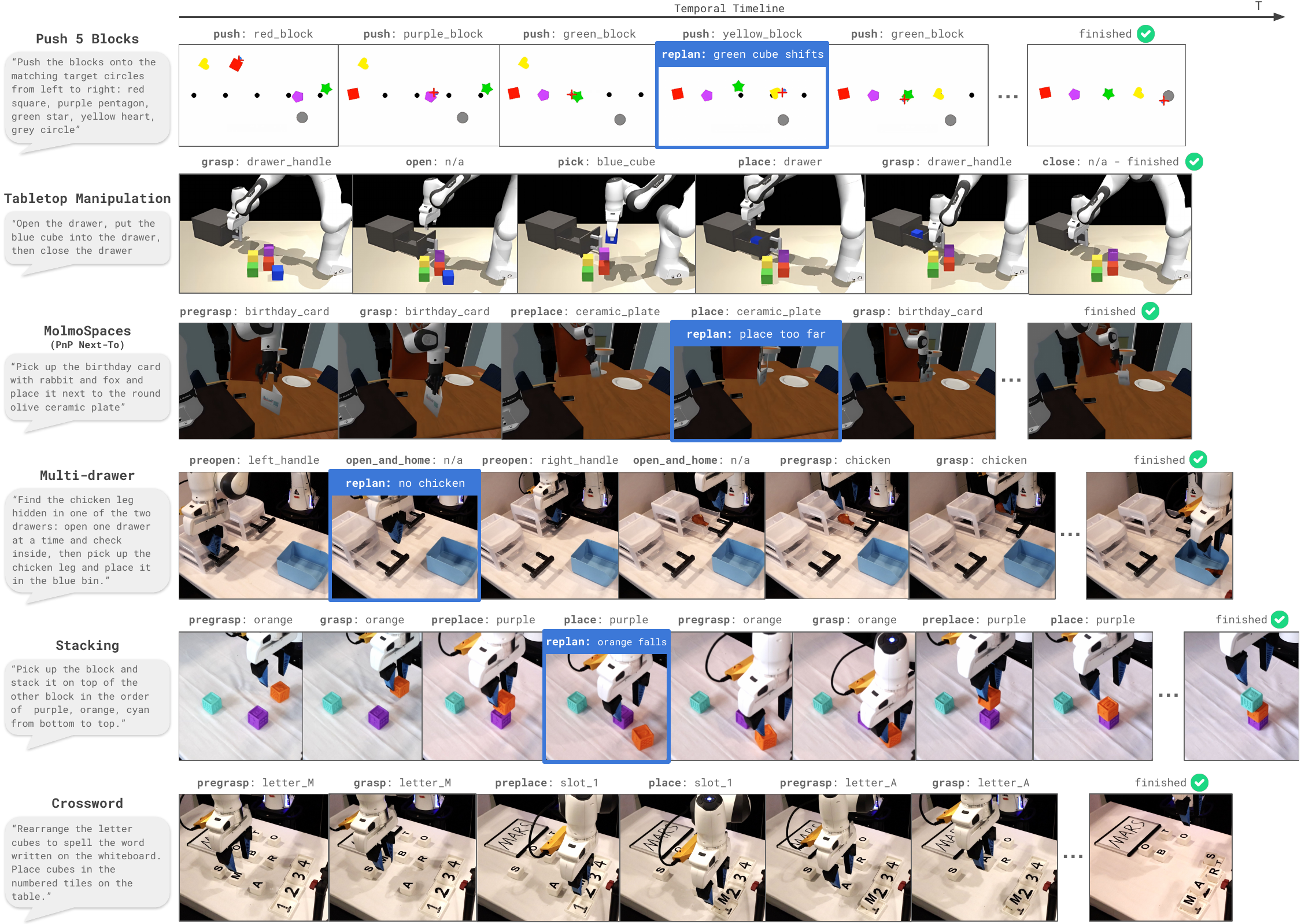}
  \caption{\textbf{Task instructions and phase execution.}
  Each row pairs an instruction with selected observations from execution.
  Bold labels name atomic phases, and the text after each colon identifies the point target; \texttt{n/a} denotes a phase without point conditioning.
  Blue boxes mark replanning after a detected failure or uncertain task state, such as a fallen cube in Stacking.
  The checker uses phase-specific completion criteria to continue execution, advance the phase, or request a new plan.
  Ellipses omit repeated or previously illustrated phases.}
  \label{fig:task_rollouts}
\end{figure*}

We evaluate \ours on three simulated suites spanning 2D pushing, Franka drawer manipulation, and generalist-policy pick-and-place, followed by real-robot tasks.
The tasks combine semantic and spatial ambiguity with dependent skill sequences, allowing us to study how steering and replanning support long-horizon execution.
All tasks use the same framework with task-specific phases, PDDL domains, and checker prompts.
\Cref{sec:appendix_setup} summarizes the simulated suites, and \cref{fig:task_rollouts} illustrates phase execution and replanning in simulation and on the real robot.

\subsection{Main Results}
\label{sec:exp_main}

\begin{table*}[!t]
\caption{\textbf{Main simulation results.} Success and score are percentages, reported as mean $\pm$ standard deviation across evaluations. VLA U-Net uses a scratch-trained byte-level or frozen CLIP text encoder. $\pi_{0.5}$\,+\,atomic uses phase prompts and checker-controlled transitions without points. DP variants use a scratch-trained ResNet or fully fine-tuned DINOv3/CLIP vision encoder. Grey rows denote \ours; bold marks the best mean in each column.}
\label{tab:main}
\centering
\small
\renewcommand{\arraystretch}{1.12}
\setlength{\tabcolsep}{7pt}
\begin{tabularx}{\textwidth}{l*{4}{>{\centering\arraybackslash}X}}
\toprule
& \multicolumn{2}{c}{Push 5 Blocks} & \multicolumn{2}{c}{Tabletop Manipulation} \\
\cmidrule(lr){2-3}\cmidrule(l){4-5}
\textbf{Method} & Success & Score & Success & Score \\
\midrule
\multicolumn{5}{l}{\textit{End-to-End Policies}} \\
VLA U-Net (scratch text enc.)  & 1.0\sd{3.0}  & 26.7\sd{5.4}  & 8.3\sd{9.0}  & 26.3\sd{8.5} \\
VLA U-Net (CLIP text enc.)     & 0.0\sd{0.0}  & 20.9\sd{7.0} & 5.0\sd{5.0}  & 20.6\sd{6.2} \\
$\pi_{0.5}$ (LoRA)             & 8.0\sd{7.5}  & 47.3\sd{5.2}  & 0.0\sd{0.0}  & 12.8\sd{5.1} \\
$\pi_{0.5}$ (full FT)          & 1.0\sd{3.0}  & 29.6\sd{5.3}  & 40.0\sd{14.1} & 60.5\sd{12.6} \\
\midrule
\multicolumn{5}{l}{\textit{Atomic Policy}} \\
\rowcolor{gray!15}
$\pi_{0.5}$\,+\,atomic         & 26.0\sd{13.6} & 68.3\sd{12.8} & 16.3\sd{14.1} & 43.4\sd{15.0} \\
\rowcolor{gray!15}
DP + ResNet                       & 71.0\sd{15.8} & 95.3\sd{3.9}  & 48.8\sd{25.7} & 68.8\sd{16.7} \\
\rowcolor{gray!15}
DP + DINOv3                       & 66.7\sd{11.9} & 93.5\sd{5.3}  & \textbf{64.0}\sd{15.0} & \textbf{82.3}\sd{8.5} \\
\rowcolor{gray!15}
DP + CLIP                         & \textbf{78.7}\sd{11.4} & \textbf{97.9}\sd{0.8} & \textbf{64.0}\sd{9.2} & 80.2\sd{6.4} \\
\bottomrule
\end{tabularx}
\end{table*}

Push 5 Blocks requires a circular pusher in a 2D physics simulator to move five blocks of different shapes and colors to their instructed target circles.
The task combines instruction following with precise object-target grounding across successive pushes.
Tabletop Manipulation requires a Franka Panda in MuJoCo to open a drawer, place the instructed cube (one of five randomly positioned colored cubes) inside, and close it.
Successful execution depends on tracking state changes across coupled skills and recovering when an earlier interaction fails.
The first two rows of \cref{fig:task_rollouts} illustrate how point targets change between pushes and how drawer opening and closing proceed without point conditioning.

By default, the planner and checker use \texttt{gemini-2.5-flash} and the grounder is MolmoPoint-8B.
We report mean $\pm$ standard deviation over up to ten evaluations, each on ten fixed held-out episodes.
We evaluate diffusion-policy (DP) instantiations of \ours with fully fine-tuned CLIP and DINOv3 encoders and a scratch-trained ResNet, alongside the following language-conditioned configurations:

(1) VLA U-Net: a language-conditioned counterpart using the same denoiser and ResNet, with either a learned byte-level or frozen CLIP text encoder.

(2) $\pi_{0.5}$: fine-tuned with task prompts using LoRA or full fine-tuning.

(3) $\pi_{0.5}$\,+\,atomic: a language-steered \ours variant that runs the fully fine-tuned $\pi_{0.5}$ with phase prompts and checker-controlled transitions, without point conditioning.

The DP instantiations of \ours outperform the end-to-end baselines on both tasks (\cref{tab:main}).
With the CLIP encoder, \ours reaches 78.7\% success on Push 5 Blocks and 64.0\% on Tabletop, compared with 8.0\% and 40.0\% for the best end-to-end baselines, respectively.
The language-conditioned VLA U-Net rarely completes either task, supporting the use of explicit phase and target conditioning in this setting.
With full fine-tuning, $\pi_{0.5}$ reaches 40.0\% success on Tabletop but only 1.0\% on Push 5 Blocks.
One possible explanation is that its pretrained manipulation priors transfer more readily to tabletop interaction than to the out-of-distribution 2D pusher.
$\pi_{0.5}$\,+\,atomic improves Push 5 Blocks success from 1.0\% to 26.0\%, showing that \ours can improve performance with the same VLA backbone.
On Tabletop, however, success drops from 40.0\% to 16.3\%, possibly reflecting difficulty following phase-specific prompts across dependent manipulation steps without explicit point guidance.
All encoders perform well on Push 5 Blocks, whereas pretrained ViTs outperform the scratch-trained ResNet by roughly 15 percentage points on Tabletop, suggesting a greater role for visual representation quality in complex manipulation.
\Cref{sec:exp_enc_vlm} further examines vision-encoder adaptation and sensitivity to the planner/checker VLM.

\subsection{Generalist Policies on MolmoSpaces}
\label{sec:exp_molmobot}

\begin{table}[t]
\caption{\textbf{MolmoSpaces PnP Next-To success (\%).} Fine-tuned results average the last 10 checkpoint evaluations ending at 16k training steps, with 20 rollouts per checkpoint (200 rollouts total). Zero-shot results use 200 rollouts per model. Grey denotes \ours.}
\label{tab:molmobot_final}
\centering
\small
\renewcommand{\arraystretch}{1.12}
\setlength{\tabcolsep}{4pt}
\begin{tabularx}{\columnwidth}{>{\raggedright\arraybackslash}Xc}
\toprule
\textbf{Method} & Success (\%) \\
\midrule
$\pi_{0.5}$ zero-shot & 3.0 \\
$\pi_{0.5}$ FT & 7.5 \\
MolmoBot zero-shot & 37.0 \\
MolmoBot FT & 43.5 \\
MolmoBot FT\,+\,Point & 27.5 \\
\midrule
\rowcolor{gray!15}
MolmoBot FT\,+\,Point\,+\,\ours & 48.5 \\
\bottomrule
\end{tabularx}
\end{table}

PnP Next-To in the MolmoSpaces~\cite{molmospaces2026} benchmark requires a Franka arm to pick up a named object and place it next to a named receptacle across diverse household scenes, including kitchens, dining areas, and living spaces (\cref{fig:task_rollouts}, third row).
Following the benchmark protocol, an episode succeeds if the moved object is placed within 5\,cm of the target object, and each episode is limited to 40 seconds.
The diverse scenes require the robot to ground object referents and spatial relations before composing the manipulation phases.
Among the simulation tasks reported in MolmoBot~\cite{molmobot2026}, PnP Next-To is the most challenging, with its best variant reaching only 28.3\% oracle success and 22.6\% final success.

We fine-tune MolmoBot~\cite{molmobot2026} on phase-labeled MolmoBot-Data and evaluate on MolmoSpaces-Bench; \cref{sec:appendix_sim} specifies the dataset and benchmark configurations.
The planner and checker use \texttt{gemini-3.8-flash}.
We evaluate four configurations based on the same MolmoBot backbone. The released DROID~\cite{khazatsky2024droid} checkpoint serves as the zero-shot baseline. The fine-tuned baselines use either the task prompt alone (FT) or the task prompt with one \texttt{<points>} tag for the pickup object and one for the receptacle, both fixed at the first frame (FT\,+\,Point). The \ours configuration uses phase prompts, per-phase points, and the VLM planning and checking loop (FT\,+\,Point\,+\,\ours).
We also evaluate $\pi_{0.5}$ zero-shot and with task-specific fine-tuning.
Fine-tuned checkpoints share 20 fixed benchmark episodes and evaluation seeds across methods. Both zero-shot models share 200 fixed benchmark episodes.

At 16k fine-tuning steps, \ours reaches 48.5\% moving-average success, compared with 43.5\% for MolmoBot FT, 27.5\% for FT\,+\,Point, and 37.0\% for zero-shot MolmoBot (\cref{tab:molmobot_final}; learning curves in \cref{sec:appendix_curves}). These results use 200 rollouts per configuration; \ours also outperforms MolmoBot FT at every reported training step.
By comparison, $\pi_{0.5}$ achieves 3.0\% zero-shot success; its fine-tuned variant reaches 7.5\% at 16k steps, with a peak moving average of 8.0\%.
With both prompt points fixed throughout the task, FT\,+\,Point must infer the active target, whereas \ours specifies it for each phase and finishes 21.0 percentage points higher at 16k steps.
The contrast with static point prompts suggests that spatial guidance benefits from being aligned with the active phase.
This comparison evaluates the combined effect of phase-specific guidance, planning, and checking on the same backbone.

\subsection{Ablation of Points and Replanning}
\label{sec:exp_ablation}

\begin{table*}[!t]
\caption{\textbf{Point-conditioning and replanning ablations.} The grey reference row uses \ours with fully fine-tuned CLIP diffusion policies. Removing replanning preserves checker-controlled transitions; removing points omits the U-Net's spatial input. The no-point Push 5 Blocks variant also omits planning and checking. Metrics follow \cref{tab:main}; bold marks the best mean.}
\label{tab:ablation}
\centering
\small
\renewcommand{\arraystretch}{1.12}
\setlength{\tabcolsep}{6pt}
\begin{tabularx}{\textwidth}{l*{6}{>{\centering\arraybackslash}X}}
\toprule
& & & \multicolumn{2}{c}{Push 5 Blocks} & \multicolumn{2}{c}{Tabletop Manipulation} \\
\cmidrule(lr){4-5}\cmidrule(l){6-7}
\textbf{Variant} & Replanning & Point & Success & Score & Success & Score \\
\midrule
\rowcolor{gray!15}
\ours                   & \cmark & \cmark & \textbf{78.7}\sd{11.4} & \textbf{97.9}\sd{0.8} & \textbf{64.0}\sd{9.2} & \textbf{80.2}\sd{6.4} \\
\quad w/o replanning    & \xmark & \cmark & 64.4\sd{10.2} & 86.9\sd{4.4} & 17.8\sd{24.4} & 40.6\sd{25.2} \\
\quad w/o points        & \cmark & \xmark & 0.0\sd{0.0} & 14.8\sd{4.1} & 4.4\sd{6.8} & 29.4\sd{5.0} \\
\bottomrule
\end{tabularx}
\end{table*}

We ablate replanning and point conditioning in the CLIP-based \ours (\cref{sec:exp_main}).
Without replanning, the checker still advances phases but cannot invoke the planner for recovery.
Without points, phase policies receive only images and proprioception.
Tabletop retains phase selection by the planner and checker, isolating spatial guidance within each phase.
Push 5 Blocks instead uses one policy throughout, without planning or checking; its comparison therefore also changes the execution loop.

Without point conditioning (\cref{tab:ablation}), success falls to 0.0\% on Push 5 Blocks and 4.4\% on Tabletop.
On Tabletop, spatial guidance improves execution even when high-level phase selection is retained.
Replanning has a larger effect on Tabletop, where phases are strongly dependent. A failed grasp or a partially opened drawer can leave later phases in an unsuitable state, and success falls from 64.0\% to 17.8\% without replanning. The phases in Push 5 Blocks are more independent, and the variant without replanning retains 64.4\% success, about 14 percentage points below the full \ours result of 78.7\%.
Together, the Tabletop ablations show why phase selection alone is insufficient: a skill still needs its intended target, and the remaining sequence must reflect the state actually reached.
Grounding and replanning address these complementary requirements.

\subsection{Real-Robot Experiments}
\label{sec:exp_real}

\begin{table*}[!t]
\caption{\textbf{Real-robot results under ID and OOD conditions.} Success is the percentage of trials with task score 1; task score measures the fraction of completed steps, scaled to 100. Task scores report mean $\pm$ sample standard deviation across 10 trials per method, task, and condition. Crossword OOD pools the two evaluated words. Grey rows denote \ours; bold marks the highest mean within each condition, including ties.}
\label{tab:real_world}
\centering
\small
\renewcommand{\arraystretch}{1.12}
\setlength{\tabcolsep}{3.5pt}
\begin{tabularx}{\textwidth}{l*{6}{>{\centering\arraybackslash}X}}
\toprule
& \multicolumn{2}{c}{Multi-drawer} & \multicolumn{2}{c}{Stacking} & \multicolumn{2}{c}{Crossword} \\
\cmidrule(lr){2-3}\cmidrule(lr){4-5}\cmidrule(l){6-7}
\textbf{Method} & Success & Task Score & Success & Task Score & Success & Task Score \\
\midrule
\multicolumn{7}{l}{\textit{In-Distribution (ID)}} \\
\rowcolor{gray!15}
Atomic + DP & 90.0 & 95.0\sd{15.8} & 20.0 & 62.5\sd{28.3} & \textbf{60.0} & \textbf{91.3}\sd{13.2} \\
\rowcolor{gray!15}
Atomic + MolmoBot & 80.0 & 89.2\sd{24.8} & 70.0 & 96.3\sd{6.0} & 50.0 & 82.5\sd{24.4} \\
\rowcolor{gray!15}
Atomic + $\pi_{0.5}$ & \textbf{100.0} & \textbf{100.0}\sd{0.0} & \textbf{90.0} & \textbf{98.8}\sd{4.0} & \textbf{60.0} & 77.5\sd{29.9} \\
\addlinespace[2pt]
MolmoBot (fine-tuned) & 80.0 & 91.3\sd{18.7} & 50.0 & 50.0\sd{52.7} & 0.0 & 30.0\sd{26.5} \\
$\pi_{0.5}$ (fine-tuned) & \textbf{100.0} & \textbf{100.0}\sd{0.0} & \textbf{90.0} & \textbf{98.8}\sd{4.0} & 10.0 & 65.0\sd{23.4} \\
\midrule
\multicolumn{7}{l}{\textit{Out-of-Distribution (OOD)}} \\
\rowcolor{gray!15}
Atomic + DP & \textbf{90.0} & \textbf{96.3}\sd{11.9} & 10.0 & 56.3\sd{23.0} & \textbf{50.0} & \textbf{90.0}\sd{12.9} \\
\rowcolor{gray!15}
Atomic + MolmoBot & 50.0 & 77.5\sd{26.1} & 30.0 & \textbf{85.0}\sd{16.5} & 30.0 & 72.5\sd{26.9} \\
\rowcolor{gray!15}
Atomic + $\pi_{0.5}$ & 50.0 & 83.3\sd{19.5} & \textbf{80.0} & \textbf{85.0}\sd{33.7} & 40.0 & 60.0\sd{41.2} \\
\addlinespace[2pt]
MolmoBot (fine-tuned) & 10.0 & 70.1\sd{13.0} & 40.0 & 62.5\sd{45.6} & 0.0 & 0.0\sd{0.0} \\
$\pi_{0.5}$ (fine-tuned) & 80.0 & 88.8\sd{25.3} & 70.0 & 70.0\sd{48.3} & 0.0 & 0.0\sd{0.0} \\
\bottomrule
\end{tabularx}
\end{table*}

We evaluate three real-robot tasks (\cref{fig:task_rollouts}).
In Multi-drawer, the robot searches two drawers for a chicken leg and places it in a bin. If the checker finds no target in the first drawer, it triggers replanning to search the other.
Stacking requires identifying and stacking colored cubes in the bottom-to-top order specified by the prompt.
If a cube falls, subsequent placements may no longer be valid, so the system must detect the failure and revise the plan from the observed state. Crossword requires reading the word on a whiteboard and placing letter blocks into numbered slots in the correct order.
The target word is specified visually rather than named in the prompt, requiring the planner to infer the letter sequence from the observation.
Together, these tasks couple ambiguous or underspecified goals with dependent manipulation phases, testing whether scene interpretation and execution feedback sustain progress over long horizons.

We use a Franka Panda robot with a Franka HAND gripper and external and wrist cameras.
Hardware and demonstration-collection details are provided in \cref{sec:appendix_real}.

We instantiate \ours with three policy backbones.
DP uses a separate point-conditioned diffusion policy for each
phase; MolmoBot receives phase prompts and point coordinates;
and $\pi_{0.5}$ receives phase prompts without point conditioning.
The end-to-end baselines fine-tune MolmoBot and $\pi_{0.5}$ on
whole demonstrations with task instructions, without phase
decomposition or point inputs.
Training epochs and optimization steps are matched, and all methods use their final checkpoints.
The planner and checker use \texttt{gpt-6-astra}, and MolmoPoint
provides spatial grounding.
The checker is queried every 128 control steps, and a phase
exceeding 256 steps triggers replanning.

Each method is evaluated on 20 trials per task (10 in-distribution and 10 out-of-distribution).
OOD conditions change cube colors, retrieval objects, and the word or visual cues used for Crossword; \cref{sec:appendix_real} records the evaluation conditions.
Task score measures the fraction of correctly completed
grasp/place steps, and success requires a score of 1.
We report task scores as percentages with sample standard deviations
across trials.

\Cref{tab:real_world} summarizes the results;
\cref{sec:appendix_trials} visualizes individual trials.
Under ID conditions, fine-tuned $\pi_{0.5}$ matches
Atomic + $\pi_{0.5}$ on Multi-drawer and Stacking,
but both baselines fall behind on Crossword.
Under OOD Crossword conditions, their rollouts frequently
follow the demonstrated word or an incorrect letter sequence
rather than the requested target.

The real-robot results expose a gap between executing learned
motions and adapting them to a changed goal.
On OOD Crossword, both end-to-end baselines obtain zero task
score, while Atomic + MolmoBot and Atomic + $\pi_{0.5}$
reach 72.5 and 60.0, respectively.
Thus, the same policy backbones become more responsive to
the requested goal when guided through explicit phases.
The improvement with Atomic + $\pi_{0.5}$, which receives
no points, shows that language-based phase guidance and
closed-loop execution alone can recover substantial
performance.
Atomic + MolmoBot combines this guidance with spatial targets;
the matched-backbone comparisons are detailed in
\cref{sec:appendix_trials}.

The failure analysis reveals a corresponding shift in
what limits performance.
On Crossword, incorrect objects, order, or placement
locations account for 36 of 39 unsuccessful baseline
trials, whereas grasp and place errors account for
25 of 31 unsuccessful \ours trials
(\cref{sec:appendix_failures}).
Explicit phase guidance therefore reduces failures in
selecting and sequencing actions, leaving physical execution
as the dominant source of remaining errors.
This distinction also explains why stronger task steering
does not guarantee success on every task:
Atomic + DP performs well on Crossword but struggles with
the precise block alignment required for Stacking.
High-level guidance can specify and retry the correct action,
but its completion still depends on the precision of the
low-level policy.

\section{Discussion}
\label{sec:discussion}

Tabletop ablations support distinct roles for target grounding and plan revision: points improve execution with phase selection retained, and replanning improves execution with checker-controlled transitions retained (\cref{tab:ablation,tab:main}).
On Crossword, fine-tuned generalist policies often reproduce the demonstrated word or follow an incorrect sequence when the visual goal changes.
\ours improves task progress, while its remaining failures predominantly involve grasping and placement (\cref{sec:exp_real,sec:appendix_failures}).
The generalist-policy comparisons assess the combined effect of phase-conditioned training, steering, and execution feedback.

Several limitations remain. The phase set, PDDL domain, and checker prompts are specified manually for each task family, and demonstrations require phase and point annotations.
Execution depends on the accuracy of the VLM and pointing model, whose inference also adds latency.
Zero-shot \texttt{gpt-6-astra} outputs end-effector displacements from camera images (\cref{sec:exp_astra}).
Direct control reaches 98.0\% success on Push 5 Blocks, compared with 86.3\% for \ours, whereas \ours reaches 60.0\% success and a score of 80.0 on Tabletop, compared with 52.0\% and 69.5.
Learned skills thus matter most for contact-rich phases that require precise and timely control, while planar pushing from a static view is within reach of direct VLM control.
Future work includes automatic phase discovery and runtime selection between direct VLM control and learned skills.

\section{Conclusion}
\label{sec:conclusion}

We presented \ours, connecting symbolic planning to learned skills through a shared steering and verification interface.
Phase-specific language and 2D image points support both specialized policies and shared action models.
Experiments in 2D pushing, tabletop manipulation, and pick-and-place demonstrate the benefits of explicit steering and feedback.
Real-robot results show improved task following and illustrate recovery in complex, long-horizon tasks with ambiguous or underspecified goals.

\clearpage
\section*{LLM Usage Statement}
We used LLM-based assistants, Claude Code and Codex, in preparing this work.
They helped polish the manuscript text, the figures, and the visualizations.
They also assisted with part of the coding in our research.
For this code, we specified the functional requirements, the implementation workflow, and the method details.
The assistants helped only with the basic low-level implementation.
We reviewed all AI-assisted content and take full responsibility for this paper.

{
    \small
    \bibliographystyle{ieeenat_fullname}
    \bibliography{main}
}

\clearpage
\appendix
\section*{Appendix}

This appendix documents the implementation (\cref{sec:appendix_method}), experimental protocols (\cref{sec:appendix_setup}), additional comparisons (\cref{sec:appendix_analysis}), and learning curves and trial outcomes (\cref{sec:appendix_visualizations}).

\section{Method Implementation Details}
\label{sec:appendix_method}
\label{sec:appendix}

The following implementation details instantiate the common execution loop introduced in \cref{sec:method}.

\subsection{Planning and Execution}
\label{sec:appendix_execution}

Each PDDL operator specifies an atomic phase through its preconditions and effects.
For the drawer task, the operators are \texttt{grasp-handle}, \texttt{open-drawer}, \texttt{pick-cube}, \texttt{place-in-drawer}, and \texttt{close-drawer}; the pushing task uses a single \texttt{push-to-slot} operator.
The VLM returns the current ground predicates and task goal in a fixed format.
A breadth-first search over ground states returns the shortest operator sequence to the goal.
The domains have at most five operators, and the solver enforces their preconditions and effects during search.
Replanning repeats state estimation and search from the current observation, so already-satisfied goals need not be executed again.

Every $C$ control steps, the checker receives the phase description together with the current primary-camera frame, wrist-camera frame, and gripper opening.
The prompt lists the visual evidence required for successful completion of each phase.
The checker returns \texttt{done}, \texttt{not\_done}, or \texttt{replan}, together with a one-sentence justification.
A phase timeout after $T$ steps also triggers replanning.
\Cref{alg:loop} gives the point-steered instantiation of this loop; for phases with $g_k=\varnothing$, grounding and point conditioning are omitted.
The no-replanning ablation in \cref{sec:exp_ablation} retains checker-controlled phase transitions but disables plan revision.

Push 5 Blocks is the exception.
Instead of the full frame, the checker receives a crop centered on the current target and judges whether the target is fully covered by the instructed block.
We use this input because the VLM often fails to locate the target in the full image, especially when a block partially occludes it during pushing.

\begin{algorithm}[H]
\caption{Closed-loop execution of \ours with point-steered skills.}
\label{alg:loop}
\small
\begin{algorithmic}[1]
\Require instruction $\ell$, phase policies $\{\pi_s\}_{s\in\mathcal{S}}$
\State $(P_1,\dots,P_K) \gets \textsc{Planner}(\ell, I_t)$;\ \ $k \gets 1$
\While{$k \le K$}
  \State $p_k \gets \textsc{Grounder}(g_k, I_t)$ \Comment{every $R$ steps}
  \State execute $A_t \sim \pi_{s_k}(\,\cdot \mid o_t, p_k)$
  \State $v \gets \textsc{Checker}(P_k, o_t)$ \Comment{every $C$ steps}
  \If{$v = \texttt{done}$}
    \State $k \gets k+1$
  \ElsIf{$v = \texttt{replan}$ \textbf{or} steps in $P_k > T$}
    \State $(P_1,\dots,P_K) \gets \textsc{Planner}(\ell, I_t)$;\ \ $k \gets 1$
  \EndIf
\EndWhile
\end{algorithmic}
\end{algorithm}

\subsection{Grounding and Prompt Interfaces}
\label{sec:appendix_interfaces}

The pointing model returns target locations in pixel coordinates.
We divide the horizontal and vertical coordinates by the primary-camera image width and height to obtain $p_k \in [0,1]^2$.
Grounding runs at the start of each target-dependent phase and every $R$ control steps thereafter.
For policies that also accept language, the phase instruction accompanies the updated point.

Each phase has a fixed prompt template, such as ``grasp the \{object\} and lift it'' for \texttt{grasp}.
The template is filled with the target identified for the active phase.
The $\pi_{0.5}$ implementation uses these phase-specific language prompts without points.
For the MolmoSpaces experiments, MolmoBot is fine-tuned on phase-labeled demonstrations from MolmoBot-Data, with an image point for each target-dependent phase.
Points are encoded using its native \texttt{<points>} tag with integer coordinates in $[0,1000]$.
All implementations use the planning and verification procedure described in \cref{sec:appendix_execution}.

\subsection{Point-Conditioned Skill Policies}
\label{sec:method_policy}

In the specialized skill bank (\cref{fig:overview}), each phase uses a diffusion policy~\cite{chi2023diffusion} that generates an action sequence $A_t = (a_t, \dots, a_{t+H-1})$ with horizon $H$. The first $H_a$ actions are executed before the policy is queried again.
The conditioning input is
\begin{equation}
  c_t = \big[\,\phi(I^{1}_{t-T_o+1:t}),\ \dots,\ \phi(I^{M}_{t-T_o+1:t}),\ q_{t-T_o+1:t},\ \psi(p_k)\,\big],
  \label{eq:cond}
\end{equation}
where $\phi$ is the vision encoder shared across camera views and applied to the latest $T_o$ frames, while $\psi$ is a three-layer MLP that maps the two point coordinates to a 128-dimensional feature.
The default encoder $\phi$ is CLIP ViT-B/32~\cite{radford2021clip}. We fully fine-tune it using one-tenth the learning rate of the rest of the network. \Cref{sec:exp_main,sec:exp_enc_vlm} compare vision backbones and fine-tuning strategies.
The denoiser $\epsilon_\theta$ is the 1D convolutional U-Net of Diffusion Policy~\cite{chi2023diffusion} and is conditioned on $c_t$.
For phases without a separate target, such as pulling an already-grasped drawer open, the point feature $\psi(p_k)$ is omitted.

\begin{table*}[!t]
\caption{\textbf{Task suites, atomic phases, and scoring.} Phases are listed in execution order. For the real-robot tasks, the task score is the fraction of correctly completed grasp and place steps (\cref{eq:task_score}), and success requires all steps to be completed.}
\label{tab:benchmarks}
\newcommand{\tph}[1]{{\footnotesize\texttt{#1}}}
\centering
\small
\setlength{\tabcolsep}{4pt}
\renewcommand{\arraystretch}{1.1}
\begin{tabular}{@{}>{\raggedright}p{3.0cm}>{\raggedright}p{7.2cm}>{\raggedright\arraybackslash}p{5.6cm}@{}}
\toprule
\textbf{Task} & \textbf{Atomic Phases} & \textbf{Success / Score} \\
\midrule
\multicolumn{3}{@{}l}{\textit{Simulation}} \\
Push 5 Blocks & \tph{push\_block} & all blocks on their targets / mean coverage \\
Tabletop Manipulation & \tph{grasp}, \tph{open}, \tph{pick}, \tph{place}, \tph{close} & cube in the closed drawer / fraction of 4 subtasks \\
MolmoSpaces (PnP Next-To) & \tph{pregrasp}, \tph{grasp}, \tph{preplace}, \tph{place}, \tph{done} & object next to the receptacle / n/a \\
\addlinespace[2pt]
\multicolumn{3}{@{}l}{\textit{Real robot}} \\
Multi-drawer & \tph{preopen}, \tph{open\_and\_home}, \tph{pregrasp}, \tph{grasp}, \tph{preplace}, \tph{place} & target object placed in the bin / fraction of grasp and place steps \\
Stacking & \tph{pregrasp}, \tph{grasp}, \tph{preplace}, \tph{place} & cubes stacked in the specified order / fraction of grasp and place steps \\
Crossword & \tph{pregrasp}, \tph{grasp}, \tph{preplace}, \tph{place} & letters placed in the numbered slots in order / fraction of grasp and place steps \\
\bottomrule
\end{tabular}
\end{table*}

Demonstrations are divided into contiguous segments using their phase labels, and each segment is assigned to the corresponding policy $\pi_s$.
For a target-dependent phase, every frame also includes the target point in the primary-camera image.
Each policy uses the standard noise-prediction objective and DDPM noise schedule~\cite{ho2020ddpm}, with noise levels sampled uniformly from the $N$ diffusion steps.
All phase policies use the same architecture and training hyperparameters.


\subsection{Wrist-Camera Pointing}
\label{sec:appendix_wrist}

Real-robot point-conditioned policies also receive a target
point in the moving wrist-camera view.
Unlike the fixed primary-camera point, this point must be
updated as the end effector moves.
Image-based tracking proved unreliable: SAM~2 drifted to
similar cubes or gripper fingers and could not consistently
isolate individual Crossword tiles.
We instead maintain a 3D target $\mathbf{x}_k$ in the robot
base frame and project it into the wrist image:
\begin{equation}
  w_t = \Pi\!\big(E(q_t)^{-1}\,\mathbf{x}_k\big),
  \qquad E(q_t) = F(q_t)\,E_{\mathrm{cam}},
  \label{eq:wrist_point}
\end{equation}
where $F(q_t)$ is the flange forward kinematics,
$E_{\mathrm{cam}}$ is the calibrated camera-to-flange
transform, and $\Pi$ is pinhole projection normalized
by image size.
A geometric visibility flag $\nu_t$ indicates whether
the target is in front of the camera and within the image.
Annotation, training, and deployment use the same geometry.

We estimate the wrist-camera intrinsics and
$E_{\mathrm{cam}}$ directly from demonstrations, without
a calibration target.
For each grasp, we approximate the cube's 3D position
by the fingertip center at gripper closure and obtain
2D correspondences by tracking the cube backward through
preceding wrist frames with SAM~2~\cite{ravi2025sam}.
Minimizing reprojection error gives a median error of
8.0 pixels on the $672{\times}376$ image.
The unchanged camera mount allows this calibration
to be shared across all three tasks. We then infer 3D targets from gripper events and fingertip
poses, sharing each target between its approach and contact
phases.
Projecting these targets with \cref{eq:wrist_point} provides
per-frame wrist points without manual frame-by-frame annotation.

For Atomic + DP, the two-dimensional point input to
$\psi$ in \cref{eq:cond} is replaced by the
five-dimensional vector $[\,p_k,\ w_t,\ \nu_t\,]$,
which concatenates the primary-camera point, the wrist
point normalized to the wrist-image crop, and the
visibility flag.
The primary-camera point remains fixed within each phase,
while the wrist point uses the most recent observation.
When $\nu_t=0$, the wrist point is set to zero.
During training, we add Gaussian noise with a standard
deviation of 0.02 to visible wrist coordinates and mask
the wrist point and visibility flag with probability 0.15.
Atomic + MolmoBot receives the visible wrist point as
an additional \texttt{<points>} tag on the current
wrist image.

During deployment, MolmoPoint-8B is queried on the wrist
image at the start of each target-dependent phase,
during the pause for primary-camera grounding.
The predicted pixel is back-projected onto a plane at
the phase's target height, estimated from median contact
heights in the demonstrations and adjusted for stack
level during placement.
This intersection initializes $\mathbf{x}_k$.
We then update $w_t$ at every control step using
\cref{eq:wrist_point}, without image processing,
including during target occlusion.
Asynchronous wrist queries refine the estimate every
16 control steps, or every 8 steps until two consistent
predictions confirm the target.
Each prediction is back-projected using the robot pose
at image capture; predictions on gripper fingers,
identified by a SAM~2 mask, are rejected.
At each policy query, the updated wrist point and
visibility flag are supplied using the same encoding
and masking rules as in training.

\subsection{Real-Robot Policy Training}
\label{sec:appendix_rw_training}

All policies are trained separately for each task on the demonstrations described in \cref{sec:appendix_real}, and each method uses the same training budget across the three tasks.
For Atomic + DP, each phase policy uses the architecture of \cref{sec:method_policy} and is trained for 500 epochs with a batch size of 256 and a cosine learning-rate schedule.
For Atomic + MolmoBot, we initialize from MolmoBot-DROID, train the action expert for 8{,}000 steps with the language model frozen, and use a batch size of 64.
For Atomic + $\pi_{0.5}$, we fully fine-tune $\pi_{0.5}$-DROID in joint-position space for 20{,}000 steps with a batch size of 32 and 15-step action chunks at 15\,Hz.
The fine-tuned MolmoBot and $\pi_{0.5}$ baselines follow the same recipes and training budgets as their \ours counterparts, with whole-task instructions in place of phase prompts and points.
We set these budgets from pilot runs on a separate sorting task, where they offered a good balance between training time and policy performance.
For every method and task, we deploy the final checkpoint and use it for all 20 evaluation trials without further checkpoint selection.

\section{Experimental Protocols}
\label{sec:appendix_setup}

\subsection{Simulation Task Suites}
\label{sec:appendix_sim}

\Cref{tab:benchmarks} summarizes the atomic phases and scoring criteria for the simulated and real-robot tasks.
The main text describes their task requirements and evaluation procedures in \cref{sec:exp_main,sec:exp_molmobot}.
The phase sets and checker inputs vary with the task, while the planning and execution procedure is shared.

For MolmoSpaces, training uses the phase-labeled \texttt{FrankaPickAndPlaceNextToOmniCamConfig} demonstrations from MolmoBot-Data, and evaluation uses \texttt{FrankaPickandPlaceNextToHardBench} from MolmoSpaces-Bench.

For each fine-tuned configuration, we average success over the last 10 checkpoint evaluations ending at 16k training steps.
Each checkpoint is evaluated on the same 20 benchmark episodes, giving 200 rollout executions per configuration.
Evaluation seeds are shared across methods and checkpoints.
This aggregation summarizes repeated execution over a fixed episode set across the final training window.
Each zero-shot model is evaluated at a single checkpoint on the same fixed set of 200 benchmark episodes.
The total rollout budget is therefore matched across configurations, while episode coverage and checkpoint allocation differ.

\subsection{Real-Robot Setup and Evaluation Conditions}
\label{sec:appendix_real}

We evaluate on a Franka Panda robot with a Franka HAND gripper, controlled through DROID+ using absolute joint-position targets. A static side-view ZED 2i camera and a wrist-mounted ZED mini camera provide visual observations. Demonstrations are collected through SpaceMouse teleoperation at approximately 14--15\,Hz. The base datasets contain 99 demonstrations for Multi-drawer, 171 for Stacking, and 148 for Crossword. Each demonstration records a complete task episode and is subsequently annotated with phase labels and target points.

Training budgets are matched in both epochs and optimization steps across the compared methods.
We use the final saved checkpoint of each trained policy throughout real-robot evaluation, including both ID and OOD conditions.

Initial cube positions are randomized for Stacking and Crossword; in Multi-drawer, the target is placed in the left and right drawers for five trials each. OOD evaluation introduces unseen cube colors in Stacking, grapes and a white bowl in Multi-drawer, and changed word, object, and visual-cue configurations in Crossword. 

The ID condition uses large letter cubes and a whiteboard
displaying \texttt{MARS}.
The OOD evaluation includes two conditions.
For \texttt{ROBO}, we change the word on the whiteboard to
\texttt{ROBO} while retaining the large-letter setup.
No large-letter demonstrations spell \texttt{ROBO}.
However, the training data include \texttt{ROBOT} demonstrations
with smaller letter cubes and a smaller numbered-slot board,
so the letters and their relative order are not entirely unseen.

For \texttt{COKE}, we replace the whiteboard cue with a Coke
drink on the table and use the instruction:
``Rearrange the letter cubes to spell the object on the table.
Place cubes in the numbered tiles on the table.''
Thus, \texttt{COKE} combines new target letters with
a change in how the goal is specified visually.
Results for both conditions are pooled into the 10 OOD
Crossword trials per method.

Each real-robot trial is divided into $K_i$ scored manipulation phases, and the evaluator records which phases the robot completes correctly.
For trial $i$ of a given method, task, and condition, the task score is
\begin{equation}
  s_i = \frac{1}{K_i}\sum_{k=1}^{K_i} c_{i,k}, \qquad c_{i,k} \in \{0, 1\},
  \label{eq:task_score}
\end{equation}
where $c_{i,k} = 1$ if phase $k$ is completed correctly and $c_{i,k} = 0$ otherwise.

\section{Additional Comparisons}
\label{sec:appendix_analysis}

These analyses complement the steering and replanning ablations in \cref{sec:exp_ablation}.
We first vary the vision encoder and planner/checker, then compare learned skill execution with direct VLM control to examine how performance depends on the task's interaction requirements.

\subsection{Vision Encoder and Planner/Checker VLM}
\label{sec:exp_enc_vlm}

\begin{table*}[!t]
\caption{\textbf{Vision-encoder and planner/checker ablations.} Vision-encoder experiments compare full and partial fine-tuning with \texttt{gemini-2.5-flash} as both planner and checker. VLM experiments use the same prompts and a fully fine-tuned CLIP vision encoder. Grey marks the reference configuration in each group; bold marks the best mean within each group. Metrics follow \cref{tab:main}.}
\label{tab:enc_vlm}
\centering
\small
\renewcommand{\arraystretch}{1.12}
\setlength{\tabcolsep}{6pt}
\begin{tabularx}{\textwidth}{l*{4}{>{\centering\arraybackslash}X}}
\toprule
& \multicolumn{2}{c}{Push 5 Blocks} & \multicolumn{2}{c}{Tabletop Manipulation} \\
\cmidrule(lr){2-3}\cmidrule(l){4-5}
\textbf{Variant} & Success & Score & Success & Score \\
\midrule
\multicolumn{5}{l}{\textit{Vision Encoder}} \\
\rowcolor{gray!15}
CLIP, full FT     & \textbf{78.7}\sd{11.4} & \textbf{97.9}\sd{0.8} & \textbf{64.0}\sd{9.2} & 80.2\sd{6.4} \\
CLIP, partial FT  & 76.7\sd{12.4} & 96.6\sd{2.1} & 40.0\sd{17.6} & 67.5\sd{11.8} \\
DINOv3, full FT   & 66.7\sd{11.9} & 93.5\sd{5.3} & \textbf{64.0}\sd{15.0} & \textbf{82.3}\sd{8.5} \\
DINOv3, partial FT & 52.2\sd{13.1} & 93.3\sd{3.9} & 27.5\sd{31.1} & 55.0\sd{21.6} \\
\midrule
\multicolumn{5}{l}{\textit{Planner/Checker VLM}} \\
\rowcolor{gray!15}
gemini-2.5-flash  & 78.7\sd{11.4} & 97.9\sd{0.8} & 64.0\sd{9.2} & 80.2\sd{6.4} \\
gemini-3.8-flash  & \textbf{83.8}\sd{9.7} & \textbf{98.7}\sd{0.7} & 48.0\sd{16.0} & 76.1\sd{9.8} \\
gpt-6-astra       & 68.1\sd{19.1} & 97.6\sd{1.6} & \textbf{65.0}\sd{11.2} & \textbf{81.9}\sd{8.2} \\
claude-opus-5.5   & 70.8\sd{19.7} & 97.5\sd{1.4} & 50.0\sd{0.0} & 72.2\sd{7.0} \\
Qwen3-VL-235B     & 47.7\sd{23.4} & 91.6\sd{3.9} & 8.6\sd{8.3} & 40.7\sd{5.5} \\
\bottomrule
\end{tabularx}
\end{table*}

We vary the vision encoder and the VLM while keeping the rest of the system fixed.
For the vision encoder, we use officially released base-sized checkpoints from two model families, CLIP ViT-B/32 and DINOv3 ViT-B/16, and compare full and partial fine-tuning within each backbone. Partial fine-tuning freezes the lower eight transformer blocks and updates the top four.
To study sensitivity to the planner/checker, we compare frontier, lower-cost, and open-weight VLMs: \texttt{gemini-3.8-flash}, \texttt{gpt-6-astra}, \texttt{claude-opus-5.5}, \texttt{gemini-2.5-flash}, and \texttt{Qwen3-VL-235B}. All models use the same prompts.

Full fine-tuning improves both encoders on both tasks (\cref{tab:enc_vlm}). On Tabletop, it improves CLIP from 40.0\% to 64.0\% and DINOv3 from 27.5\% to 64.0\%, gains of 24.0 and 36.5 percentage points, respectively. This suggests that full adaptation better captures manipulation-specific visual cues.
Among the VLMs, \texttt{gemini-3.8-flash} performs best on Push 5 Blocks at 83.8\%, while \texttt{gpt-6-astra} performs best on Tabletop at 65.0\%. \texttt{Qwen3-VL-235B} performs worst on both tasks, particularly on Tabletop, which may reflect difficulty with the spatial and state reasoning required for planning and completion checking. The other four models are broadly competitive. The lower-cost \texttt{gemini-2.5-flash} reaches 78.7\% on Push 5 Blocks and 64.0\% on Tabletop, remaining competitive with the other models. The VLM ranking depends on the task.

\subsection{Zero-Shot Direct VLM Control}
\label{sec:exp_astra}

\begin{table}[!ht]
\caption{\textbf{Direct VLM control versus learned skill execution.} The direct controller is \texttt{gpt-6-astra}. Success and score are reported as mean $\pm$ std (\%) across five 10-episode blocks from the same 50 episodes. Grey denotes \ours; bold marks the higher mean. Both methods achieve high task scores on Push 5 Blocks; \ours achieves higher mean success and score with lower variation on simulated Tabletop manipulation.}
\label{tab:astra}
\centering
\small
\renewcommand{\arraystretch}{1.12}
\setlength{\tabcolsep}{2.5pt}
\begin{tabularx}{\columnwidth}{l*{4}{>{\centering\arraybackslash}X}}
\toprule
& \multicolumn{2}{c}{Push 5 Blocks} & \multicolumn{2}{c}{Tabletop} \\
\cmidrule(lr){2-3}\cmidrule(l){4-5}
\textbf{Method} & Success & Score & Success & Score \\
\midrule
gpt-6-astra   & \textbf{98.0}\sd{4.0} & \textbf{99.2}\sd{1.6} & 52.0\sd{16.0} & 69.5\sd{11.4} \\
\rowcolor{gray!15}
\ours         & 86.3\sd{10.9} & 95.5\sd{3.2} & \textbf{60.0}\sd{6.3} & \textbf{80.0}\sd{4.2} \\
\bottomrule
\end{tabularx}
\end{table}

We also test whether a frontier VLM can control the robot directly without task-specific policy training.
\texttt{gpt-6-astra} receives the current camera images, end-effector state, and a history of its previous actions. At each query, it outputs up to four relative end-effector displacements for the simulator to execute. \texttt{gpt-6-astra} is used zero-shot. \ours uses the CLIP encoder and its saved policy checkpoints, while selecting the best planner/checker VLM for each task from \cref{tab:enc_vlm}: \texttt{gemini-3.8-flash} on Push 5 Blocks and \texttt{gpt-6-astra} on Tabletop.
Both methods are evaluated on the same 50 seeded episodes.

Push 5 Blocks is a simplified planar benchmark with a static top-down view and limited dexterity demands.
Direct control with \texttt{gpt-6-astra} is highly effective, reaching 98.0\% success compared with 86.3\% for \ours (\cref{tab:astra}).
Both approaches achieve high task scores, with \ours reaching 95.5 compared with 99.2 for direct control.

Tabletop is a more realistic simulation of articulated manipulation, requiring precise grasps and timely control at 10\,Hz across dependent actions.
Here, \ours reaches 60.0\% success and a score of 80.0, compared with 52.0\% and 69.5 for direct VLM control, with less variation across episode blocks.
Both configurations use \texttt{gpt-6-astra} on Tabletop.
The gains therefore support coupling its task-level reasoning and verification with trained, point-steered skills for precise local execution.

\section{Learning Curves and Trial Outcomes}
\label{sec:appendix_visualizations}

\subsection{MolmoSpaces Learning Curves}
\label{sec:appendix_curves}

\begin{figure}[H]
  \centering
  \includegraphics[width=\linewidth]{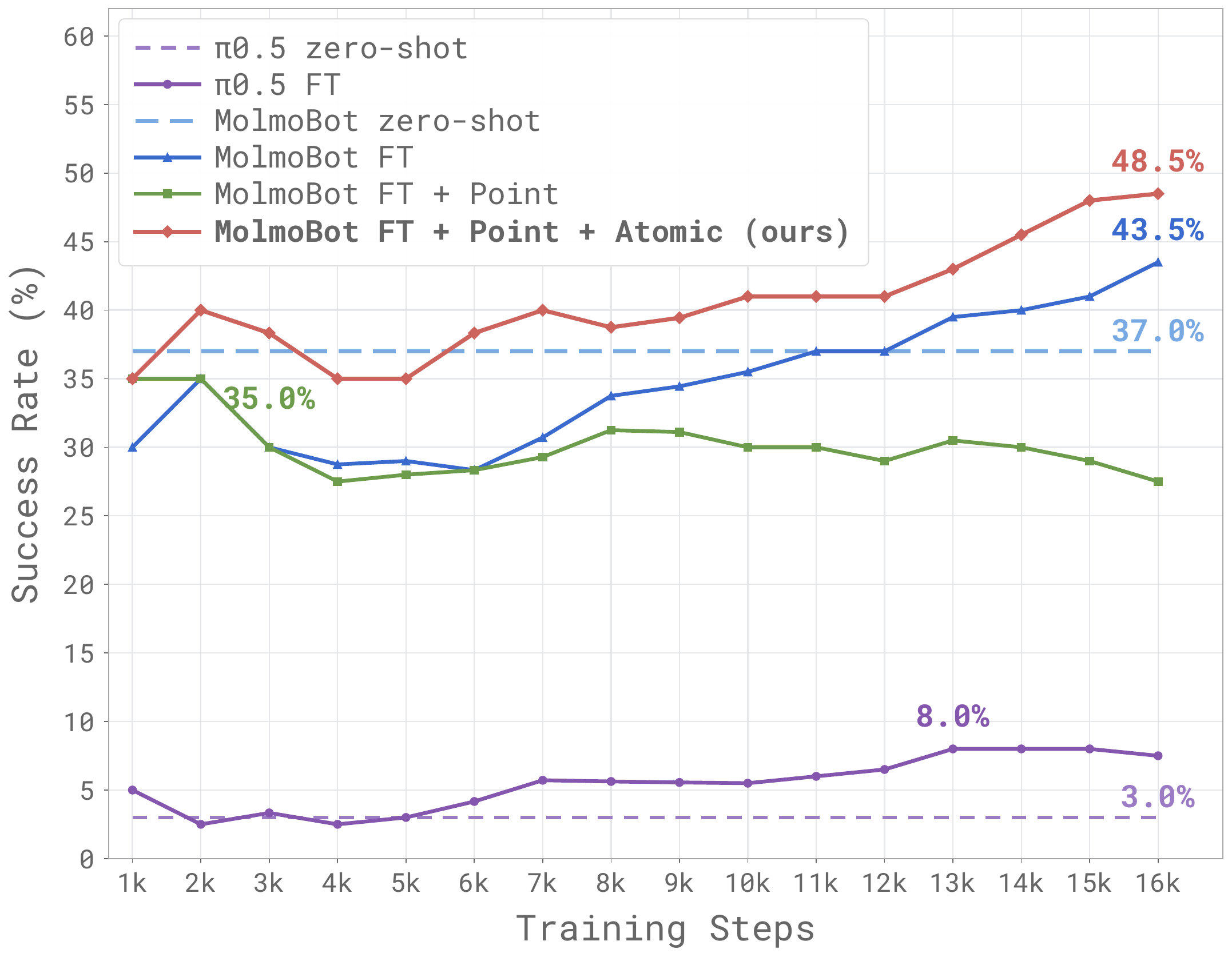}
  \caption{\textbf{Learning curves on MolmoSpaces PnP Next-To.} MolmoBot and $\pi_{0.5}$ are evaluated during task-specific fine-tuning.
  Solid curves show success averaged over the ten most recent checkpoint evaluations.
  For the first nine checkpoints, the average uses all checkpoints evaluated so far.
  Dashed horizontal lines indicate each model's zero-shot success rate.}
  \label{fig:molmobot_curve}
\end{figure}

\Cref{fig:molmobot_curve} shows how evaluation success evolves during fine-tuning for the configurations in \cref{sec:exp_molmobot}.
The curves complement the final-window comparison in \cref{tab:molmobot_final} by showing the trajectory of performance across training.

\newpage
\subsection{Real-Robot Trial Outcomes}
\label{sec:appendix_trials}

\begin{figure*}[!t]
  \centering
  \includegraphics[width=\textwidth]{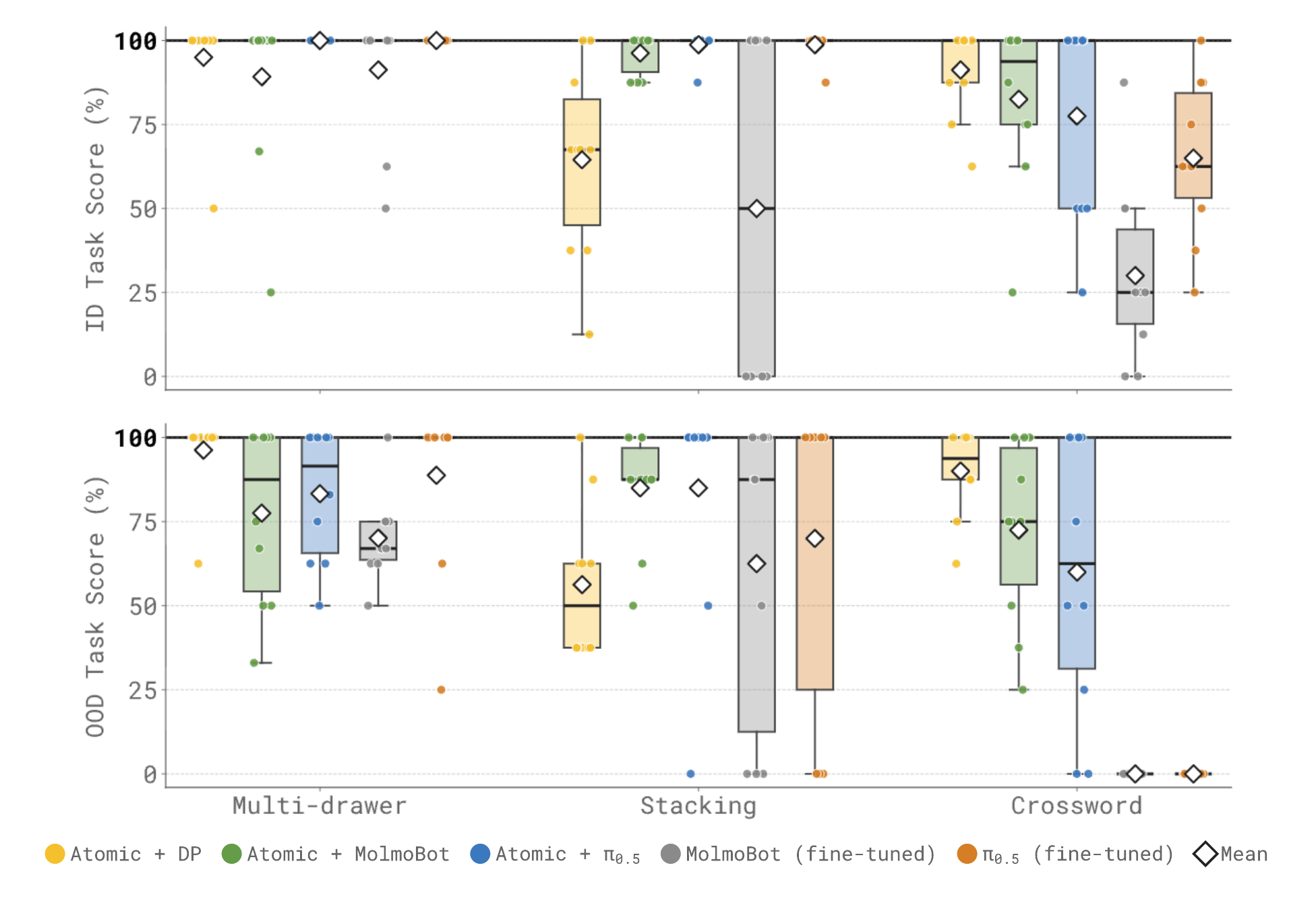}
    \caption{\textbf{Real-robot trial outcomes.} Top: in-distribution (ID); bottom: out-of-distribution (OOD). Boxes show the interquartile range (IQR), dark lines mark
    medians, and whiskers extend to the outermost observations
    within 1.5 IQR of the box edges.
    White diamonds mark means, and dots show individual trials
    ($n=10$ per method, task, and condition).
    Tasks are arranged by increasing language-grounding and
    compositional demands, from Multi-drawer to Stacking to
    Crossword. The language-only end-to-end baselines show wide trial-to-trial
    variation on Stacking and poor performance on Crossword,
    suggesting limited steerability as language-conditioning
    demands increase.
    Atomic + MolmoBot improves over fine-tuned MolmoBot on
    OOD Stacking and Crossword, supporting the combined benefit
    of point guidance and atomic-phase execution.
    Atomic + $\pi_{0.5}$ also improves on these tasks without
    point inputs, showing that high-level planning and
    checker-controlled execution can improve OOD performance
    through language-based phase guidance alone.}
  \label{fig:real_world}
\end{figure*}

\Cref{fig:real_world} complements \cref{tab:real_world} by showing individual trial scores and their variation across tasks and conditions.
These outcomes complement the analysis in \cref{sec:exp_real} of task interpretation, skill dependencies, and recovery across the real-robot tasks.

Relative to their corresponding fine-tuned baselines,
Atomic + MolmoBot and Atomic + $\pi_{0.5}$ improve OOD
Crossword task score by 72.5 and 60.0 percentage points,
respectively.
The corresponding gains are 22.5 and 15.0 points on
Stacking, and $+7.4$ and $-5.5$ points on Multi-drawer.
Thus, the largest gains occur on Crossword, where execution
depends on grounding a visually specified target into
individual letter selections and an ordered sequence.
This pattern illustrates the value of grounding task intent from visual context and maintaining that intent across a dependent skill sequence.

\begin{figure*}[!t]
  \centering
  \includegraphics[width=\textwidth]{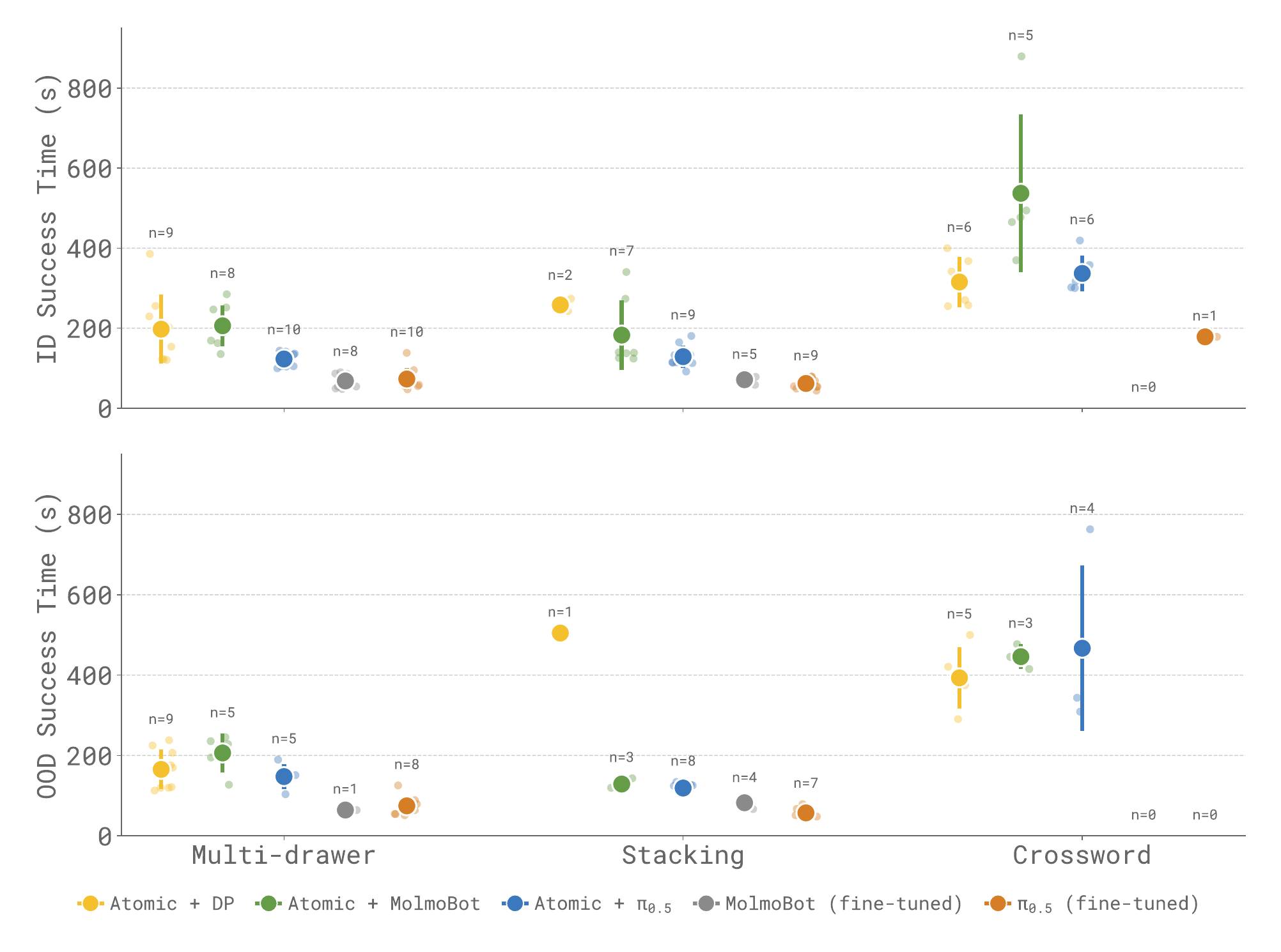}
  \caption{\textbf{Completion time of successful real-robot trials.} Top: ID; bottom: OOD. Only trials with a task score of 1 are included. Large dots show the mean completion time, vertical lines show $\pm$ one sample standard deviation, and faint dots show individual trials; $n$ is the number of successful trials for each method. Fine-tuned MolmoBot completes no Crossword trial, and fine-tuned $\pi_{0.5}$ completes a single ID Crossword trial. Completion time for \ours includes planning, verification, and replanning by the VLM, so successful \ours trials take longer than successful end-to-end trials; for example, Atomic + $\pi_{0.5}$ averages 128.6\,s on ID Stacking versus 61.6\,s for fine-tuned $\pi_{0.5}$. Means based on few successes, such as $n \le 2$, should be interpreted with caution.}
  \label{fig:success_time}
\end{figure*}

\Cref{fig:success_time} reports the completion time of successful trials.
The decomposition adds overhead from VLM planning, phase verification, and replanning.
Among the \ours variants, Atomic + $\pi_{0.5}$ is the fastest on both tasks in ID and OOD conditions.
On Crossword, this overhead accompanies a substantially higher success rate, whereas the fine-tuned baselines complete only one trial in total.

\subsection{Real-Robot Failure Modes}
\label{sec:appendix_failures}

\begin{figure*}[!t]
  \centering
  \includegraphics[width=\textwidth]{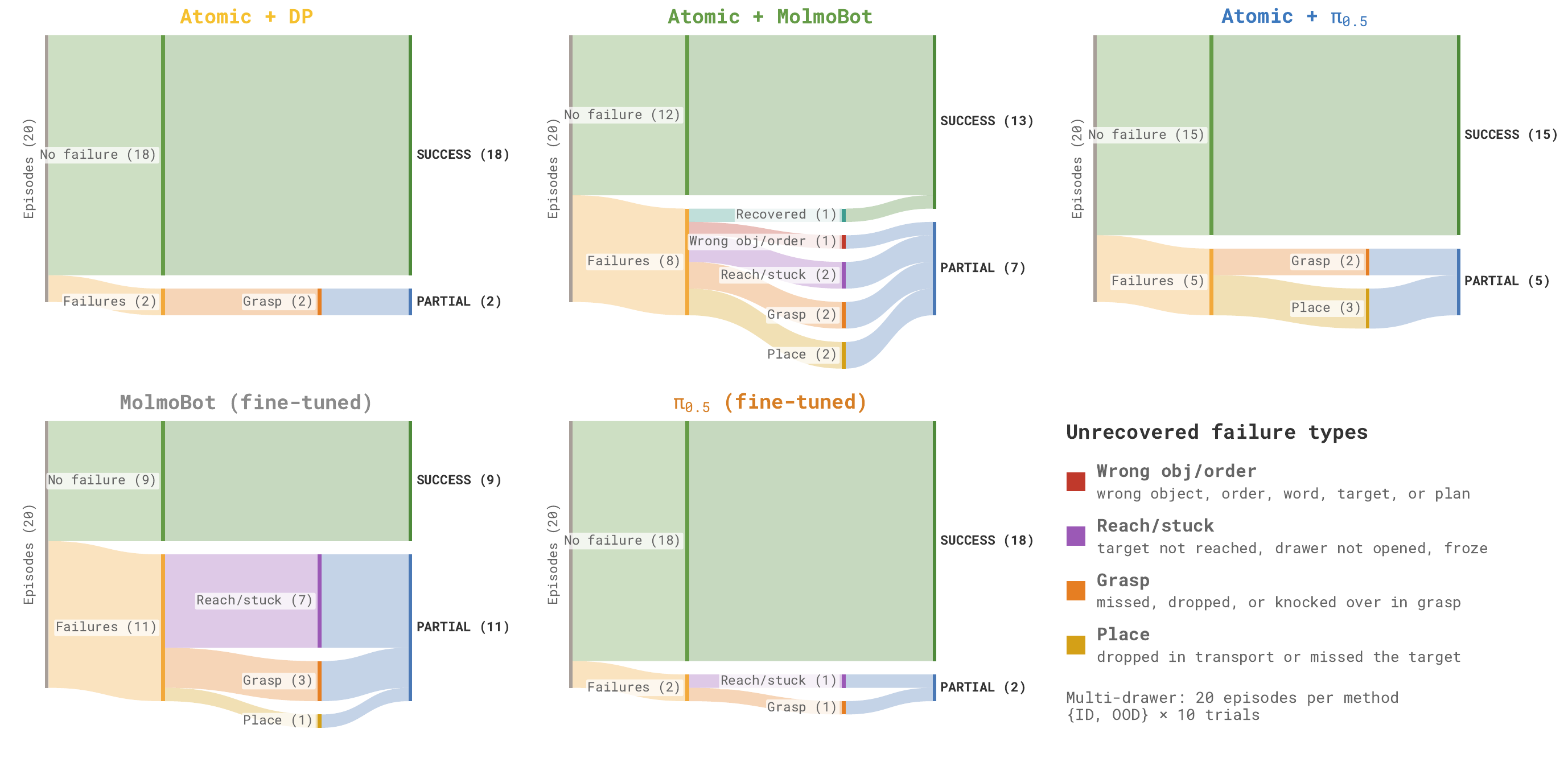}
  \caption{\textbf{Failure modes on Multi-drawer.} Each panel traces the 20 trials of one method (10 ID and
10 OOD) from execution to outcome.
Trials either proceed without a recorded failure or encounter
an error that is recovered through replanning or remains
unresolved.
Outcomes are SUCCESS (task score 1), PARTIAL
($0 <$ task score $< 1$), and FAIL (task score 0).
Failure types are wrong object or order (incorrect object,
sequence, word, target recognition, or plan), reach or stuck
(target or receptacle not reached, drawer not opened,
or robot motion stalls), grasp, and place.
Recovery is counted only when explicitly recorded in the
evaluation notes.
No method produces a zero-score trial on Multi-drawer.
The \ours variants fail mainly through grasp and place
errors, with one planning error for Atomic + MolmoBot.
For fine-tuned MolmoBot, 7 of its 11 unsuccessful trials
involve failure to reach the OOD receptacle during placement:
the policy fails to guide the grasped object to the white
bowl, suggesting difficulty grounding the unfamiliar
receptacle's location.}
  \label{fig:failure_md}
\end{figure*}

\begin{figure*}[!t]
  \centering
  \includegraphics[width=\textwidth]{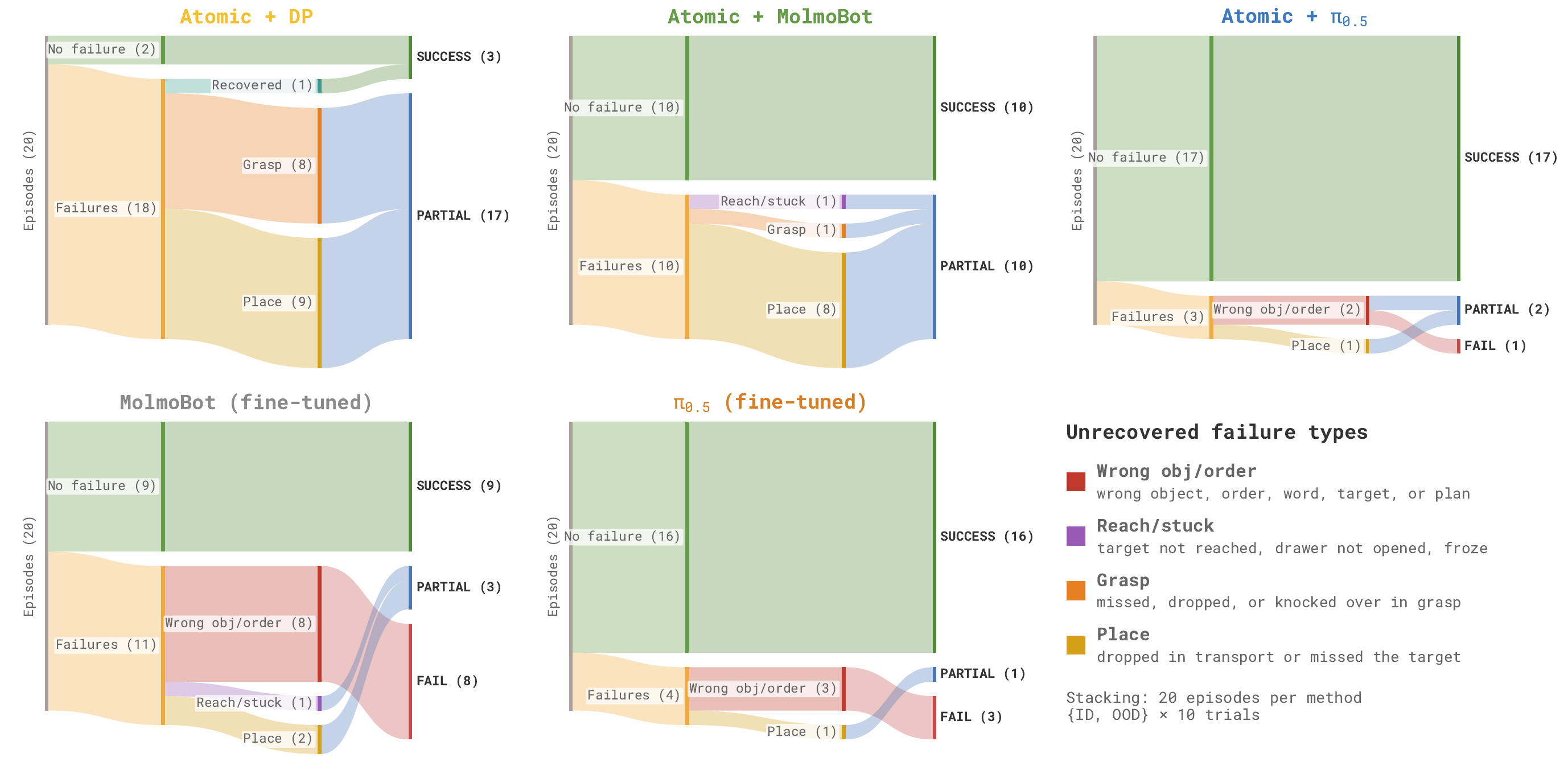}
  \caption{\textbf{Failure modes on Stacking.} Panels and failure types follow \cref{fig:failure_md}.
All 17 unrecovered failures of Atomic + DP and 9 of 10
of Atomic + MolmoBot are grasp or place errors, while
Atomic + $\pi_{0.5}$ fails in only 3 of 20 trials.
Atomic + DP struggles with the fine-grained alignment
required for stacking, suggesting that its low-level
visuomotor policy has not learned sufficient placement
precision from the available demonstrations.
High-level replanning can retry these actions but cannot
ensure accurate execution.
In contrast, wrong object or order accounts for 8 of 11
failed trials of fine-tuned MolmoBot and 3 of 4 of
fine-tuned $\pi_{0.5}$, which together produce 11
zero-score trials.}
  \label{fig:failure_stacking}
\end{figure*}

\begin{figure*}[!t]
  \centering
  \includegraphics[width=\textwidth]{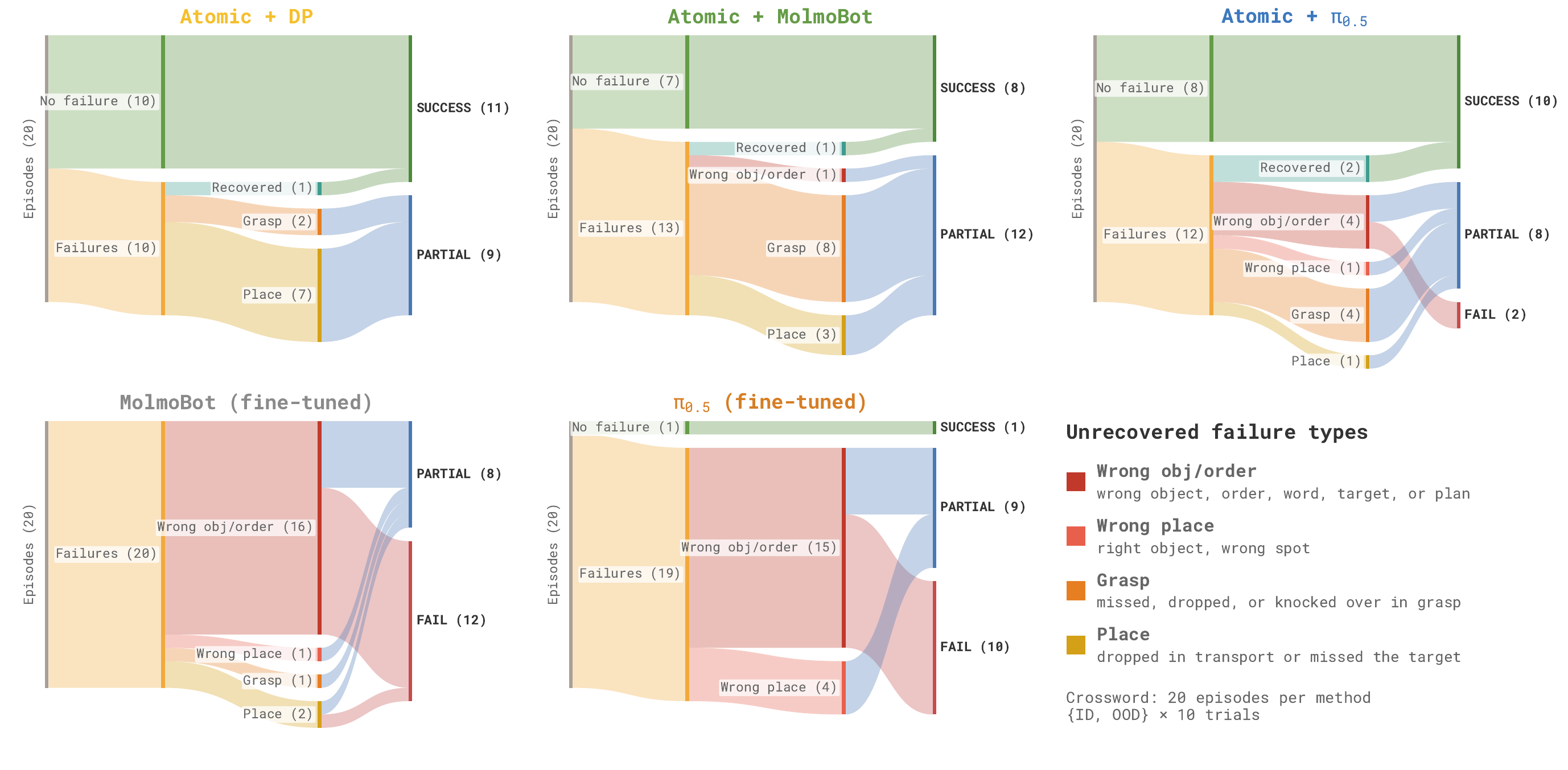}
  \caption{\textbf{Failure modes on Crossword.} Panels and failure types follow \cref{fig:failure_md}.
Wrong object, order, or place accounts for 36 of 39
unsuccessful trials of the fine-tuned baselines, whose
rollouts often follow the training word or misorder letters.
In contrast, 25 of 31 unrecovered failures across the
\ours variants are grasp or place errors, and four trials
succeed after recovery.
Together with \cref{tab:real_world}, these results support
the benefit of explicit guidance and closed-loop planning
for both ID execution and OOD generalization:
Atomic + MolmoBot raises task score from 30.0 to 82.5
on ID trials and from 0.0 to 72.5 on OOD trials.
Atomic + DP achieves the highest task scores in both
conditions (91.3 ID and 90.0 OOD), showing that a
point-guided policy can perform well on a task with
strong language-grounding and sequencing demands.}
  \label{fig:failure_crossword}
\end{figure*}

\Cref{fig:failure_md,fig:failure_stacking,fig:failure_crossword}
trace execution and outcomes across 300 real-robot rollouts:
20 trials per method and task (10 ID and 10 OOD).
Trials that do not complete the task are classified by
the error that ends execution.
We distinguish unsuccessful trials, which include partial
completion, from zero-score failures.

Multi-drawer. No method receives a zero task score, indicating that all
trials complete at least some manipulation steps.
The remaining failures of the \ours variants are mainly
grasp and place error. For fine-tuned MolmoBot, 7 of 11 unsuccessful trials involve
failure to reach the white bowl during OOD placement.
The policy grasps the object but fails to transport it
to the unfamiliar receptacle, suggesting a limitation in
target grounding rather than grasp execution alone.

Stacking. All 17 unsuccessful trials of Atomic + DP and 9 of 10
of Atomic + MolmoBot end in grasp or place errors.
Atomic + DP particularly struggles with the precise
alignment needed to place one cube on another.
Replanning can select a new action or retry a failed step,
but successful stacking still requires sufficient precision
from the low-level policy.

Crossword. The distinction between task-following and execution errors
is most pronounced on Crossword.
Wrong object, order, or place accounts for 36 of 39
unsuccessful baseline trials.
The policies often revert to the demonstrated word or
select an incorrect letter sequence instead of following
the visually specified target.
Across the \ours variants, 25 of 31 unsuccessful trials
instead end in grasp or place errors; the other six involve
a wrong object, order, or placement location.
Four additional trials encounter an error but succeed
after recorded recovery.

Together, task score, full-task success, and the failure
analysis provide complementary views of hierarchical
manipulation. Task score captures partial progress toward
the requested goal, while full-task success measures whether
that progress is sustained through completion.
The failure analysis further distinguishes errors in target
selection and action ordering from grasping and placement
errors, helping interpret the contributions of high-level
guidance and low-level execution.

This distinction is particularly informative on Crossword,
where explicit phase and target guidance substantially
improves task scores and shifts the dominant failure mode
from incorrect letter selection and ordering to physical
execution. On tasks with simpler goal-grounding and sequencing
demands, end-to-end policies can already follow the intended
goal, making manipulation reliability a more prominent
factor in overall performance.
For example, on OOD Stacking, Atomic + MolmoBot raises task
score from 62.5 to 85.0, despite completing three rather than
four of the ten trials. These results highlight the complementary
roles of explicit guidance in maintaining goal-directed
progress and reliable motor skills in converting that
progress into complete task execution.

\end{document}